\documentclass{article}

\usepackage[preprint, nonatbib]{nips_2018}

\usepackage{float}

\usepackage[square,numbers]{natbib}
\usepackage[utf8]{inputenc} 
\usepackage[T1]{fontenc}    
\usepackage{hyperref}       
\usepackage{url}            
\usepackage{booktabs}       
\usepackage{amsfonts}       
\usepackage{nicefrac}       
\usepackage{microtype}      
\usepackage{graphicx}
\usepackage{tabularx}
\usepackage{hyperref}

\title{Evaluation of sentence embeddings in downstream and linguistic probing tasks}

\author{
  Christian S. Perone \\
  \texttt{christian.perone@gmail.com} \\
\And
 Roberto Silveira \\
  \texttt{rsilveira79@gmail.com} \\
\And
Thomas S. Paula \\
  \texttt{tsp.thomas@gmail.com} \\
}

\begin{document}
\maketitle

\begin{abstract}
Despite the fast developmental pace of new sentence embedding methods, it is still challenging to find comprehensive evaluations of these different techniques. In the past years, we saw significant improvements in the field of sentence embeddings and especially towards the development of universal sentence encoders that could provide inductive transfer to a wide variety of downstream tasks. In this work, we perform a comprehensive evaluation of recent methods using a wide variety of downstream and linguistic feature probing tasks. We show that a simple approach using bag-of-words with a recently introduced language model for deep context-dependent word embeddings proved to yield better results in many tasks when compared to sentence encoders trained on entailment datasets. We also show, however, that we are still far away from a universal encoder that can perform consistently across several downstream tasks.
\end{abstract}

\section{Introduction}
Word embeddings are nowadays pervasive on a wide spectrum of Natural Language Processing (NLP) and Natural Language Understanding (NLU) applications. These word representations improved downstream tasks in many domains such as machine translation, syntactic parsing, text classification, and machine comprehension, among others~\cite{Camacho-Collados}. Ranging from count-based to predictive or task-based methods, in the past years, many approaches were developed to produce word embeddings, such as Neural Probabilistic Language Model~\cite{Bengio2003}, Word2Vec~\cite{mikolov2013distributed}, GloVe~\cite{pennington2014glove}, and more recently ELMo~\cite{peters2018deep}, to name a few.

Although most of the recent word embedding techniques rely on the distributional linguistic hypothesis, they differ on the assumptions of how meaning or context are modeled to produce the word embeddings. These differences between word embedding techniques can have unsuspected implications regarding their performance in downstream tasks as well as in their capacity to capture linguistic properties. Nowadays, the choice of word embeddings for particular downstream tasks is still a matter of experimentation and evaluation.

Even though word embeddings produce high-quality representations for words (or sub-words), representing large chunks of text such as sentences, paragraphs or documents is still an open research problem~\cite{conneau2017supervised}. The tantalizing idea of learning sentence representations that could achieve good performance on a wide variety of downstream tasks, also called universal sentence encoder is, of course, the major goal of many sentence embedding techniques. However, as we will see, we are still far away from a universal representation that has consistent performance on a wide range of tasks.

A common approach for sentence representations is to compute the Bag-of-Words (BoW) of the word vectors, traditionally using a simple arithmetic mean of the embeddings for the words in a sentence along the words dimension. This usually yielded limited performance, however, some recent methods demonstrated important improvements over the traditional averaging. By using weighted averages and modifying them using singular-value decomposition (SVD), the method known as \emph{smooth inverse frequency} (SIF)~\cite{arora2016simple}, proved to be a strong baseline over traditional averaging. Recently, \textit{p}-mean~\cite{DBLP:journals/corr/abs-1803-01400} also demonstrated improvements over SIF and traditional averaging by concatenating \emph{power means} of the embeddings, closing the gap with other complex sentence embedding techniques such as InferSent~\cite{conneau2017supervised}.

Other sentence embedding techniques were also developed based on encoder/decoder architectures, such as the Skip-Thought~\cite{kiros2015skip}, where the skip-gram model from Word2Vec~\cite{mikolov2013distributed} was abstracted to form a sentence level encoder that is trained on a self-supervised fashion. Recently, bi-directional LSTM models were also employed by InferSent~\cite{conneau2017supervised} on a supervised training scheme using the Stanford Natural Language
Inference (SNLI) dataset~\cite{bowman2015large} to predict entailment/contradiction. InferSent~\cite{conneau2017supervised} proved to yield much better results on a variety of downstream tasks when compared to many strong baselines or self-supervised methods such as Skip-Thought~\cite{kiros2015skip}, by leveraging strong supervision. Lately, the Universal Sentence Encoder (USE)~\cite{DBLP:journals/corr/abs-1803-11175} mixed an unsupervised task using a large corpus together with the supervised SNLI task and showed a significant improvement by leveraging the Transformer architecture~\cite{vaswani2017attention}, which is solely based on attention mechanisms, although without providing an evaluation with other baselines and previous works such as InferSent~\cite{conneau2017supervised}.

Neural Language Models can be tracked back to~\cite{Bengio2003}, and more recently deep bi-directional language models (biLM)~\cite{peters2018deep} have successfully been applied to word embeddings in order to incorporate contextual information. Very recently, \cite{radford2018unsupervised} used unsupervised generative pre-training of language models followed by discriminative fine-tunning to achieve state-of-the-art results in several NLP downstream tasks (improving 9 out of 12 tasks). The authors concluded that using language model as objective to fine-tuning helped both in model generalization and convergence. A similar approach of transfer learning using pre-trained language model was previously presented in ULMFiT paper~\cite{DBLP:journals/corr/abs-1801-06146}, with surprisingly good results even when fine-tuning using small datasets. 

Despite the fast development of a variety of recent methods for sentence embedding, there are no extensive evaluations covering recent techniques on common grounds. The developmental pace of new methods has surpassed the pace of inter-methodology evaluation. SentEval~\cite{conneau2018senteval} was recently proposed to reduce this comparison gap and the common problems associated with the evaluation of sentence embeddings, creating a common evaluation pipeline to assess the performance on different downstream tasks.

Recently, \cite{conneau2018you} introduced an evaluation method based on 10 probing tasks designed to capture linguistic properties from sentence embeddings, which was later integrated into SentEval~\cite{conneau2018senteval}. However, many recent sentence embedding techniques were not evaluated in this pipeline, such as the Universal Sentence Encoder~\cite{DBLP:journals/corr/abs-1803-11175}.

In this work, we describe an extensive evaluation of many recent sentence embedding techniques. We perform an analysis of the transferability of these embeddings to downstream tasks as well as their linguistic properties through the use of probing tasks \cite{conneau2018you}. We heavily make use of recent evaluation protocols based on SentEval \cite{conneau2018senteval}, creating a useful panorama of the current sentence embedding techniques and drawing important conclusions about their performance for different tasks.

This paper is organized as follows: In Section~\ref{sec:related-word}, we review work on both word and sentence embeddings, which are the basis of our experiments. In Section~\ref{sec:evaluation-tasks} we describe the evaluation tasks and datasets employed for the evaluations and, in Section~\ref{sec:methods}, we describe the evaluated models and the methodology used to generate the sentence embeddings. In Section~\ref{sec:experimental-results}, we describe the experimental results in downstream tasks and linguistic probing tasks. Finally, we summarize the contribution of this work in the concluding section.

\section{Related Work}
\label{sec:related-word}

Word embeddings are extensively used in state-of-the-art NLP techniques, mainly due to their ability to capture semantic and syntactic information of words using large unlabeled datasets and providing an important inductive transfer to other tasks.

There are several implementations of word embeddings in the literature. Following the pioneering work by \cite{Bengio2003} on the neural language model for distributed word representations, the seminal Word2Vec~\cite{mikolov2013distributed} is one of the first popular approaches of word embeddings based on neural networks. This type of representation is able to preserve semantic relationships between words and their context, where context is modeled by nearby words. In \cite{mikolov2013distributed}, they presented two different methods to compute word embeddings: Skip-gram (SG), which predicts context words given a target word and Continuous Bag-of-Words (CBOW), which predicts target word using a bag-of-words context. Word2Vec was later found \cite{NIPS2014_5477} to be implicitly factorizing a word-context matrix, where the cells are the pointwise mutual information (PMI) of the respective word and context pairs.

Global Vectors (GloVe)~\cite{pennington2014glove} aims to overcome some limitations of Word2Vec, focusing on the global context for learning the representations. The global context is captured by the statistics of word co-occurrences in a corpus (count-based, as opposed to the prediction-based method as in Word2Vec), while still capturing semantic and syntactic meaning as in Word2Vec.

FastText~\cite{bojanowski2016enriching} is a recent method for learning word embeddings for large datasets. It can be seen as an extension of Word2Vec that treats each word as a composition of character n-grams. The sub-word representation allows fastText to represent words more efficiently, enabling the estimation of rare and out-of-vocabulary (OOV) words. In ~\cite{joulin2016bag} the authors used fastText word representation combined with techniques such as bag of n-gram features and demonstrated that fastText obtained performance on par with deep learning methods, while being faster.

Two main challenges exist when learning high-quality representations: they should capture semantic and syntax and the different meanings the word can represent in different contexts (polysemy). To solve these two issues, Embedding from Language Models (ELMo)~\cite{peters2018deep} was recently introduced. It uses representations from a bi-directional LSTM that is trained with a language model (LM) objective on a large text dataset. ELMo~\cite{peters2018deep} representations are a function of the internal layers of the bi-directional Language Model (biLM), which provides a very rich representation about the tokens.

Like in fastText \cite{bojanowski2016enriching}, ELMo~\cite{peters2018deep} breaks the tradition of word embeddings by incorporating sub-word units, but ELMo~\cite{peters2018deep} has also some fundamental differences with previous shallow representations such as fastText or Word2Vec. In ELMo~\cite{peters2018deep}, they use a deep representation by incorporating internal representations of the LSTM network, therefore capturing the meaning and syntactical aspects of words. Since ELMo~\cite{peters2018deep} is based on a language model, each token representation is a function of the entire input sentence, which can overcome the limitations of previous word embeddings where each word is usually modeled as an average of their multiple contexts.

Through the lens of the Ludwig Wittgenstein philosophy of language \cite{wittgenstein1953philosophical}, it is clear that the ELMo~\cite{peters2018deep} embeddings are a better approximation to the idea of ``meaning is use'' \cite{wittgenstein1953philosophical}, where a word can contain a wide spectrum of different meanings depending on context, as opposed to traditional word embeddings that are not only context-independent but have a very limited definition of context.

\begin{table}
  \caption{Comparison of different sentence embeddings evaluated in this work. These embedding sizes are the final sentence embeddings (i.e. after applying BoW). Since $p$-mean~\cite{DBLP:journals/corr/abs-1803-01400} is a concatenation of other embeddings, its training method was left unspecified.}
  \label{sample-table}
  \centering
  \begin{tabular}{llc}
    \toprule
    Name    & Training method\footnotemark     & Embedding size \\
    \midrule
    ELMo (BoW, all layers, 5.5B) & Self-supervised        & 3072   \\
    ELMo (BoW, all layers, original)  & Self-supervised        & 3072   \\
    ELMo (BoW, top layer, original)  & Self-supervised        & 1024   \\
    Word2Vec (BoW, Google news)       & Self-supervised        & 300  \\
    $p$-mean (monolingual)       & --        & 3600 \\
    FastText (BoW, Common Crawl)          & Self-supervised          & 300   \\
    GloVe (BoW, Common Crawl)            & Self-supervised        & 300  \\
    USE (DAN)            & Supervised          & 512 \\
    USE (Transformer)               & Supervised          & 512 \\
    InferSent (AllNLI)          & Supervised          & 4096 \\
    Skip-Thought           & Self-supervised        & 4800 \\
    \bottomrule
  \end{tabular}

\end{table}
\footnotetext{We adopt here the term \emph{self-supervised} for some tasks, however, literature often use the term  \emph{unsupervised} as well. The authors of this work believe that the self-supervised term help to disambiguate situations that can lead to a misunderstanding of the training task.}

Although bag-of-words of word embeddings showed good performance for some tasks, it is still unclear how to properly represent the full sentence meaning. Nowadays, there is still no consensus on how to represent sentences and many studies were proposed towards that research direction.

Skip-Thought Vectors~\cite{kiros2015skip} are based on a sentence encoder that, instead of predicting the context of a word as Word2Vec, it predicts the surrounding sentences of a given sentence. It is based on encoder-decoder models, where the encoder (usually based on RNNs) maps words to a sentence vector and the decoder generates the surrounding sentences. A major advantage of Skip Thought Vectors for representing sentences when compared with a simple average of word embeddings is that order is considered during the encoding/decoding process.

InferSent~\cite{conneau2017supervised} proposes a supervised training for the sentence embeddings, contrasting with previous works such as Skip-Thought. The sentence encoders are trained using Stanford Natural Language Inference (SNLI) dataset, which consists of 570k human-generated English sentence-pairs and it is considered one of the largest high-quality labeled datasets for building sentence semantics understanding~\cite{bowman2015large}. The authors tested 7 different architectures for the sentence encoder and the best results are achieved with a bi-directional LSTM (BiLSTM) encoder.

\textit{p}-mean~\cite{DBLP:journals/corr/abs-1803-01400} emerged as a response to InferSent~\cite{conneau2017supervised} and baselines such as Sent2Vec~\cite{pagliardini2017unsupervised}. According to the authors, averaging the word embeddings and comparing with approaches such as InferSent~\cite{conneau2017supervised} can be unfair due to the difference in embedding dimensions (e.g. 300 vs 4096). \textit{p}-mean is a method that concatenates different word embeddings that represent different information such as syntactic and semantic information, resulting in a larger representation for the word embeddings. In addition, the computation of mean is based on \textit{power means}~\cite{hardy1952inequalities}.

Google recently introduced Universal Sentence Encoder~\cite{DBLP:journals/corr/abs-1803-11175}, where two different encoders were implemented. The first is the Transformer based encoder model~\cite{vaswani2017attention}, which aims for high-accuracy but has larger complexity and uses more computational resources. The second model uses a deep averaging network (DAN)~\cite{iyyer2015deep}, where embeddings for words and bi-grams are averaged together and then used as input to a deep neural network that computes the sentence embeddings.

Some other efforts in creating sentences embeddings include, but are not limited to, Doc2Vec/Paragraph2Vec~\cite{le2014distributed}, fastSent~\cite{hill2016learning} and Sent2Vec~\cite{pagliardini2017unsupervised}. In our work, we did not include these other approaches since we believe that the chosen ones are already representative of the existing ones and can enable indirect comparisons with omitted methods.

\section{Evaluation tasks}
\label{sec:evaluation-tasks}
In this section, we describe the evaluation tasks that were employed to asses the performance on downstream or linguistic probing tasks.

\subsection{Downstream tasks}
One of the main issues with both word and sentence embeddings is the evaluation procedure. One approach is to make use of such embeddings in downstream tasks, evaluating how suitable they are to different problems and which kind of semantic information they carry. Another approach is to explore the nature of the semantics by experimental methods from cognitive sciences~\cite{bakarov2018survey}.

To evaluate each method, we used the entire set of tasks and datasets available on the SentEval \cite{conneau2018senteval} evaluation framework. These tasks cover a wide range of different tasks that are suitable for general-purpose/universal sentence embeddings. These tasks can be divided into 5 groups: binary and multi-class classification, entailment and semantic relatedness, semantic textual similarity, paraphrase detection, and caption-image retrieval. Please refer to the original SentEval \cite{conneau2018senteval} article for more information about these tasks. We provide a description and sample instances of these datasets and tasks in Table \ref{table-downstream-classification-senteval} for the classification tasks and in Table \ref{table-downstream-senteval} for the semantic similarity tasks.

\begin{table}[!htb]
  \caption{Downstream classification tasks description and samples.}
  \label{table-downstream-classification-senteval}
  \centering
    \resizebox{.9\textwidth}{!}{
    \begin{tabularx}{\textwidth}{>{\centering}p{3.0cm}Xp{3.5cm}c}
    \toprule
    \textbf{Dataset}                     & \textbf{Task} & \textbf{Example} & \textbf{Output} \\
    \midrule
    Customer Reviews (CR)~\cite{hu2004mining}                & Sentiment analysis of customer products' reviews & We tried it out Christmas night and it worked great~. & Positive\\
    \midrule
    Multi-Perspective Question and Answering (MPQA)~\cite{wiebe2005annotating} & Evaluation of opinion polarity & Don't want & Negative\\
    \midrule
    Movie Reviews (MR)~\cite{pang2005seeing}                   & Sentiment analysis of movie reviews & Too slow for a younger crowd , too shallow for an older one . & Negative\\
    \midrule
    Stanford Sentiment Analysis $2$ (SST-$2$)~\cite{socher2013recursive}    & Sentiment analysis with two classes: Negative and Positive & Audrey Tautou has a knack for picking roles that magnify her [..] & Positive\\
    \midrule
    Stanford Sentiment Analysis $5$ (SST-$5$)~\cite{socher2013recursive}    & Sentiment analysis with $5$ classes, that range from $0$ (most negative) to $5$ (most positive) & Nothing about this movie works & $0$\\
    \midrule
    Subjectivity / Objectivity (SUBJ)~\cite{pang2004sentimental}    & Classify the sentence as Subjective or Objective & A movie that doesn't aim too high , but doesn't need to . & Subjective\\
    \midrule
    Text REtrieval Conference (TREC)~\cite{voorhees2000building}     & Question and answering & What are the twin cities ? & LOC:city\\
\bottomrule
    \end{tabularx}
    }
\end{table}

\begin{table}[!htb]
  \caption{Downstream semantic relatedness and textual similarity tasks descriptions and samples.}
  \label{table-downstream-senteval}
  \centering
    \resizebox{.8\textwidth}{!}{
    \begin{tabularx}{\textwidth}{>{\centering}p{2.5cm}p{2.5cm}XXc}
    \toprule
    \textbf{Dataset}                     & \textbf{Task} & \textbf{Sentence $1$} & \textbf{Sentence $2$} & \textbf{Output} \\
    \midrule
    Microsoft Common Objects in Context (COCO)~\cite{lin2014microsoft}  & Image-caption retrieval (ICR) & - & A group of people on some horses riding through the beach & Rank\\
    \midrule
    Microsoft Research Paraphrase Corpus (MRPC)~\cite{dolan2004unsupervised} & Classify whether a pair of sentences capture a paraphrase relationship & The procedure is generally performed in the second or third trimester.&  The technique is used during the second and, occasionally, third trimester of pregnancy & Paraphase\\
    \midrule
    Semantic Text Similarity (STS)~\cite{cer2017semeval}       & To measure the semantic similarity between two sentences from $0$ (not similar) to $5$ (very similar) & Liquid ammonia leak kills 15 in Shanghai & Liquid ammonia leak kills at least 15 in Shanghai & $4.6$\\
    \midrule
    Sentences Involving Compositional Knowledge Entailment (SICK-E)~\cite{marelli2014sick} & To measure semantics in terms of Entailment, Contradiction, or Neutral & A man is sitting on a chair and rubbing his eyes & There is no man sitting on a chair and rubbing his eyes & Contradiction\\
    \midrule
    Sentences Involving Compositional Knowledge Semantic Relatedness (SICK-R)~\cite{marelli2014sick} & To measure the degree of semantic relatedness between sentences from $0$ (not related) to $5$ (related) & A man is singing a song and playing the guitar & A man is opening a package that contains headphones & $1.6$ \\
    \midrule
    Stanford Natural Language Inference (SNLI)~\cite{bowman2015large} & To measure semantics in terms of Entailment, Contradiction, or Neutral & A small girl wearing a pink jacket is riding on a carousel & The carousel is moving & Entailment\\
\bottomrule
    \end{tabularx}
    }
\end{table}


\subsection{Linguistic probing tasks}
Downstream tasks are not suitable to understand what the representations are in fact capturing from the linguistic perspective. Probing tasks are classification problems that focus on simple linguistic properties of the sentences~\cite{conneau2018you}. We executed experiments using the 10 probing tasks proposed by~\cite{conneau2018you} and a summary of the tasks with examples is shown in Table~\ref{table-probing-senteval}. Each task aims to capture a different linguistic property. For instance, Coordination Invertion (CoordInv) measures whether two coordinate clauses in a sentence are inverted or not, while Past Present (Tense) aims to detect if the main verb in a given sentence is in the present or past tense. For more information about these tasks, please refer to the original article \cite{conneau2018you}.

\begin{table}[!htb]
  \caption{Linguistic probing tasks description and samples.}
  \label{table-probing-senteval}
  \centering
  \resizebox{.8\textwidth}{!}{
    \begin{tabularx}{\textwidth}{>{\centering}p{3.0cm}Xp{3.5cm}c}
    \toprule
    \textbf{Task}                       & \textbf{Description} & \textbf{Example} & \textbf{Output} \\
    \midrule
    Bigram Shift (BShift)               & Whether two words (tokens) in a sentence have been inverted & This is my Eve Christmas~. & Inverted\\
    \midrule
    Coordination Inversion (CoordInv)   & Sentences comprised of two coordinate clauses. Detect whether clauses are inverted & I returned to my work , and Lisa headed for her office . & Inverted\\
    \midrule
    Object Number (ObjNum)              & Number of the direct object in the main clause (singular and plural) & He received the 200 points~. & NNS (Plural)\\
    \midrule
    Sentence Length (SentLen)           & Predict the sentence length among $6$ classes, which are length intervals & I can 't wait to show you and Mr. Taylor . & $9-12$ words\\
    \midrule
    Semantic Odd Man Out (SOMO)         & Random noun or verb replaced in the sentence by another noun or verb. Detect whether the sentence has been modified & Tomas surmised as well~. & Changed\\
    \midrule
    Subject Number (SubjNum)            & Number of the subject in the main clause (singular and plural) & If there was ever a time to let loose , this vacation would have to be it~. & Singular\\
    \midrule
    Past Present (Tense)                & Whether the main verb in the sentence is in the past or present tense & She smiled at him , her eyes alight with love~. & Present\\
    \midrule
    Top-Constituent (TopConst)          & Classification task, where the classes are given by the $19$ most common top-constituent sequences in the corpus & Did he buy anything from Troy~? & VBD\_NP\_VP\_\\
    \midrule
    Depth of Syntactic Tree (TreeDepth) & Predict the maximum depth of the syntactic tree of the sentence & The leaves were in various of stages of life~. & $10$\\
    \midrule
    Word Content (WC)                   & Predict which of the target words (among $1000$) appear in the sentence & She eyed him skeptically~. & eyed\\
    \bottomrule
    \end{tabularx}
    }
\end{table}

\section{Methods}
\label{sec:methods}

\subsection{Experimental setup}
In this section, we describe where the pre-trained models were obtained as well as the procedures employed to evaluate each method.

\textbf{ELMo (BoW, all layers, 5.5B)}~\cite{peters2018deep}: this model was obtained from the authors' website at \url{https://allennlp.org/elmo}. According to the authors, the model was trained on a dataset with 5.5B tokens consisting of Wikipedia (1.9B) and all of the monolingual news crawl data from WMT 2008-2012 (3.6B). To evaluate this model, we used the AllenNLP framework~\cite{Gardner2017AllenNLP}. An averaging bag-of-words was employed to produce the sentence embeddings, using features from all three layers of the ELMo~\cite{peters2018deep} model. We did not employ the trainable task-specific weighting scheme described in~\cite{peters2018deep}.

\textbf{ELMo (BoW, all layers, original)}~\cite{peters2018deep}: this model was obtained from the authors website at \url{https://allennlp.org/elmo}. According to the authors, the model was trained on the 1 Billion Word Benchmark, approximately 800M tokens of news crawl data from WMT 2011. To evaluate this model, we used the AllenNLP framework~\cite{Gardner2017AllenNLP}. An averaging bag-of-words was employed to produce the sentence embeddings, using features from all three layers of the ELMo~\cite{peters2018deep} model and averaging along the word dimension. We did not employ the trainable task-specific weighting scheme described in~\cite{peters2018deep}.

\textbf{ELMo (BoW, top layer, original)}~\cite{peters2018deep}: the same model and procedure as in \emph{ELMo (BoW, all layers, original)} was employed, except that in this experiment, we used only the top layer representation from the ELMo~\cite{peters2018deep} model. As shown in~\cite{peters2018deep}, the higher-level LSTM representations capture context-dependent aspects
of meaning, while the lower level
representations capture aspects of syntax. Therefore, we split the evaluation of the top layer from the evaluation using all layers described in the previous experiment. We did not employ the trainable task-specific weighting scheme described in~\cite{peters2018deep}.

\textbf{FastText (BoW, Common Crawl)}~\cite{bojanowski2016enriching}: this model was obtained from the authors website at \url{https://fasttext.cc/docs/en/english-vectors.html}. According to the authors, this model contains 2 million word vectors trained on Common Crawl (600B tokens) dataset. A traditional bag-of-words averaging was employed to produce the sentence embedding.

\textbf{GloVe (BoW, Common Crawl)}~\cite{pennington2014glove}: this model was obtained from the authors website at \url{https://nlp.stanford.edu/projects/glove/}. According to the authors, it contains a 2.2M vocabulary and was trained on the Common Crawl (840B tokens) dataset. A traditional bag-of-words averaging was employed to produce the sentence embedding.

\textbf{Word2Vec (BoW, Google News)}~\cite{mikolov2013distributed}: this model was obtained from the authors website at \url{https://code.google.com/archive/p/word2vec/}. According to the authors, it was trained on part of the Google News dataset (about 100 billion words). A traditional bag-of-words averaging was employed to produce the sentence embedding.

\textbf{$p$-mean (monolingual)}~\cite{DBLP:journals/corr/abs-1803-01400}: this model was obtained from the authors website at \url{https://github.com/UKPLab/arxiv2018-xling-sentence-embeddings}. A TensorFlow (TF-Hub) module was employed and the sentences were all made lowercase as per authors website recommendation.

\textbf{Skip-Thought}~\cite{kiros2015skip}: this model was obtained from the authors website at \url{https://github.com/ryankiros/skip-thoughts}. The sentences were embedded according to the authors' website instructions.

\textbf{InferSent (AllNLI)}~\cite{conneau2017supervised}: this model was obtained from the authors website at \url{https://github.com/facebookresearch/InferSent}. According to the authors, it was trained on the SNLI and MultiNLI datasets. The sentences were embedded according to the authors website instructions.

\textbf{USE (DAN)}~\cite{DBLP:journals/corr/abs-1803-11175}: the Universal Sentence Encoder (USE) was obtained from the TF Hub website at \url{https://tfhub.dev/google/universal-sentence-encoder/1}. According to the TF Hub website, the model was trained with a deep averaging network (DAN) encoder~\cite{iyyer2015deep}.

\textbf{USE (Transformer)}~\cite{DBLP:journals/corr/abs-1803-11175}: the Universal Sentence Encoder (USE) was obtained from the TF Hub website at \url{https://www.tensorflow.org/hub/modules/google/universal-sentence-encoder-large/1}. According to the TF Hub website, the model was trained with a Transformer~\cite{vaswani2017attention} encoder.

\subsection{Downstream classification tasks}
As in~\cite{conneau2018senteval}, a classifier was employed on top of the sentence embeddings for the classification tasks. In this work, a Multi-Layer Perceptron (MLP) was used with a single hidden layer of 50 neurons with no dropout added, using Adam~\cite{journals/corr/KingmaB14} optimizer and a batch size of 64. We provide more information about the number of classes and validation scheme employed for each task in Table~\ref{tab:downstream-classfication-setup}.

\begin{table}[h]
\centering
\caption{Downstream classification tasks setup.}
\label{tab:downstream-classfication-setup}
    \resizebox{.7\textwidth}{!}{
\begin{tabular}{lccl}
\toprule
\textbf{Task}               & \textbf{Classifier} & \textbf{Num. Classes} & \textbf{Validation}                 \\ \toprule

\multicolumn{1}{l|}{CR}     & MLP                 & 2                     & 10-fold cross-validation (nested)   \\
\multicolumn{1}{l|}{MPQA}   & MLP                 & 2                     & 10-fold cross-validation (nested)   \\
\multicolumn{1}{l|}{MR}     & MLP                 & 2                     & 10-fold cross-validation (nested)   \\
\multicolumn{1}{l|}{SUBJ}   & MLP                 & 2                     & 10-fold cross-validation (nested)   \\
\multicolumn{1}{l|}{TREC}   & MLP                 & 6                     & 10-fold cross-validation            \\
\multicolumn{1}{l|}{MRPC}   & MLP                 & 2                     & 10-fold cross-validation            \\
\multicolumn{1}{l|}{SICK-E} & MLP                 & 3                     & Standard cross-validation (holdout) \\
\multicolumn{1}{l|}{SST-2}  & MLP                 & 2                     & Standard cross-validation (holdout) \\
\multicolumn{1}{l|}{SST-5}  & MLP                 & 5                     & Standard cross-validation (holdout) \\ \bottomrule
\end{tabular}
}
\end{table}

\subsection{Semantic relatedness and textual similarity tasks}
We used the same scheme as in~\cite{conneau2018senteval} to evaluate the semantic relatedness (SICK-R, STS Benchmark) and semantic textual similarity (STS-[12-16]). For semantic relatedness, which predicts a semantic value between 0 and 5 between two input sentences, we learn to predict the probability distribution of relatedness scores.
For the semantic textual similarity, where the goal is to assess how the cosine similarity between two sentences
correlates with a human annotation, we employed a Pearson correlation coefficient. For more information about these tasks, please refer to the SentEval~\cite{conneau2018senteval} paper.

\subsection{Information retrieval tasks}
In the caption-image retrieval task, each image and language features are jointly evaluated with the objective of ranking a collection of images in respect to a given caption (image retrieval task - \textit{text2image}) or ranking captions with respect to a given image (caption retrieval - \textit{image2text}). The dataset used to evaluate the quality of image and caption retrieval tasks in SentEval is the Microsoft COCO~\cite{lin2014microsoft}, which contains 91 common object categories present in 2,5 million labeled instances in 328k images. SentEval used 113k images from COCO dataset, each containing 5 captions. The metric used to rank caption and image retrieval in this task is recall at K (Recall@K), with K = {1, 5, 10}, and also median over 5 splits of 1k images. COCO uses a ResNet-101~\cite{he2016deep} for image embedding extraction, yielding 2048-d representation.

\subsection{Linguistic probing tasks}
For the linguistic probing tasks, a MLP was also used with a single hidden layer of 50 neurons, with no dropout added, using Adam~\cite{journals/corr/KingmaB14} optimizer with a batch size of 64, except for the Word Content (WC) probing task, as in~\cite{conneau2018you}, in which a Logistic Regression was used since it provided consistently better results.

\section{Experimental results}
\label{sec:experimental-results}

\subsection{Downstream classification tasks}
\begin{figure}[!htb]
 \includegraphics[width=\linewidth]{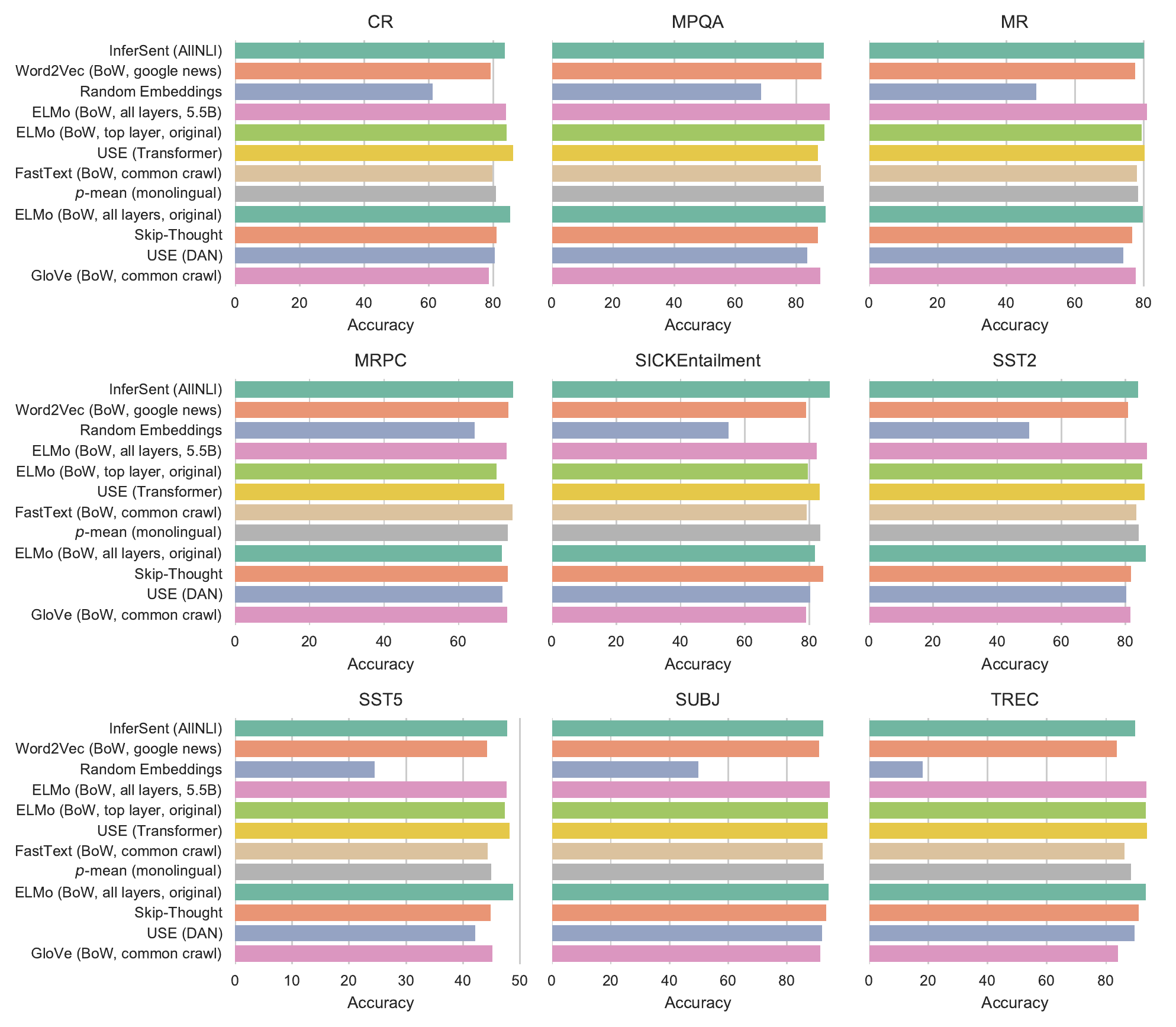}
\caption{Results for the downstream classifications tasks using a MLP. Best viewed in color.}
\label{fig:downstream-classification-tasks}
\end{figure}

In Table~\ref{tab:downstream-classification-results} we show the tabular results for the downstream classification tasks, and in Figure~\ref{fig:downstream-classification-tasks} we show a graphical comparison between the different methods. As seen in Table~\ref{tab:downstream-classification-results}, although no method had a consistent performance among all tasks, ELMo~\cite{peters2018deep} achieved best results in 5 out of 9 tasks. Even though ELMo~\cite{peters2018deep} was trained on a language model objective, it is important to note that in this experiment a bag-of-words approach was employed. Therefore, these results are quite impressive, which lead us to believe that excellent results can be obtained by integrating ELMo~\cite{peters2018deep} and the trainable task-specific weighting scheme described in~\cite{peters2018deep} into InferSent~\cite{conneau2017supervised}. 

InferSent~\cite{conneau2017supervised} achieved very good results in the paraphrase detection as well as in the SICK-E (entailment). We hypothesize that these results were due to the similarity of these tasks to the tasks were InferSent~\cite{conneau2017supervised} was trained on (SNLI and MultiNLI). As described in~\cite{conneau2017supervised}, the SICK-E can be seen as an out-domain version of the SNLI dataset.

The Universal Sentence Encoder (USE) ~\cite{DBLP:journals/corr/abs-1803-11175} model, with the Transformer encoder, also achieved good results on the product review (CR) and on the question-type (TREC) tasks. Given that the USE model was trained on SNLI as well as on web question-answer pages, it is possible that these results were also due to the similarity of these tasks to the training data employed by the USE model.

$p$-mean~\cite{DBLP:journals/corr/abs-1803-01400} also performed better on most tasks than a simple bag-of-words of GloVe, Word2Vec or fastText independently, and it is a recommended strong baseline when computational resources are limited. 

As we can see, sentence embedding methods are still far away from the idea of a universal sentence encoder that can have a broad transfer quality. Given that ELMo~\cite{peters2018deep} demonstrated excellent results on a broad set of tasks, it is clear that a proper integration of deep representation from language models can potentially improve sentence embedding methods by a significant margin and it is a promising research line.

For completeness, we also provide an evaluation using Logistic Regression instead of a MLP in Table \ref{table-downstream-logreg-senteval-result} of the appendix.

\begin{table}[!htb]
\centering
\caption{Results from downstream classification tasks results using a MLP. Values in this table are accuracies for the test set.}
\label{tab:downstream-classification-results}
\resizebox{\textwidth}{!}{%
\begin{tabular}{lccccccccc}
\toprule
\textbf{Approach}                     & \textbf{CR}    & \textbf{MPQA}  & \textbf{MR}    & \textbf{MRPC}  & \textbf{SICK-E} & \textbf{SST-2} & \textbf{SST-5} & \textbf{SUBJ}  & \textbf{TREC}  \\
\midrule
\textit{Baseline}             &       &       &       &       &        &       &       &       &       \\
\midrule
Random Embedding     & 61.16 & 68.41 & 48.75 & 64.35 & 54.94  & 49.92 & 24.48 & 49.83 & 18.00 \\
\midrule
\textit{Experiments}          &       &       &       &       &        &       &       &       &       \\
\midrule
ELMo (BoW, all layers, 5.5B) & 83.95 & \textbf{91.02} & \textbf{80.91} & 72.93 & 82.36  & \textbf{86.71} & 47.60 & \textbf{94.69} & 93.60 \\
ELMo (BoW, all layers, original)      & 85.11 & 89.55 & 79.72 & 71.65 & 81.86  & 86.33 & \textbf{48.73} & 94.32 & 93.40 \\
ELMo (BoW, top layer, original)      & 84.13 & 89.30 & 79.36 & 70.20 & 79.64  & 85.28 & 47.33 & 94.06 & 93.40 \\
Word2Vec (BoW, google news)             & 79.23 & 88.24 & 77.44 & 73.28 & 79.09  & 80.83 & 44.25 & 90.98 & 83.60 \\
$p$-mean (monolingual)               & 80.82 & 89.09 & 78.34 & 73.22 & 83.52  & 84.07 & 44.89 & 92.63 & 88.40 \\
FastText (BoW, common crawl)             & 79.63 & 87.99 & 78.03 & 74.49 & 79.28  & 83.31 & 44.34 & 92.19 & 86.20 \\
GloVe (BoW, common crawl)                & 78.67 & 87.90 & 77.63 & 73.10 & 79.01  & 81.55 & 45.16 & 91.48 & 84.00 \\
USE (DAN)         & 80.50 & 83.53 & 74.03 & 71.77 & 80.39  & 80.34 & 42.17 & 91.93 & 89.60 \\
USE (Transformer)    & \textbf{86.04} & 86.99 & 80.20 & 72.29 & 83.32  & 86.05 & 48.10 & 93.74 & \textbf{93.80} \\
InferSent (AllNLI)            & 83.58 & 89.02 & 80.02 & \textbf{74.55} & \textbf{86.44}  & 83.91 & 47.74 & 92.41 & 89.80 \\
SkipThought          & 81.03 & 87.06 & 76.60 & 73.22 & 84.33  & 81.77 & 44.80 & 93.33 & 91.00 \\
\bottomrule
\end{tabular}}
\end{table}

\subsection{Semantic relatedness and textual similarity tasks}
\begin{figure}[!htb]
 \includegraphics[width=\linewidth]{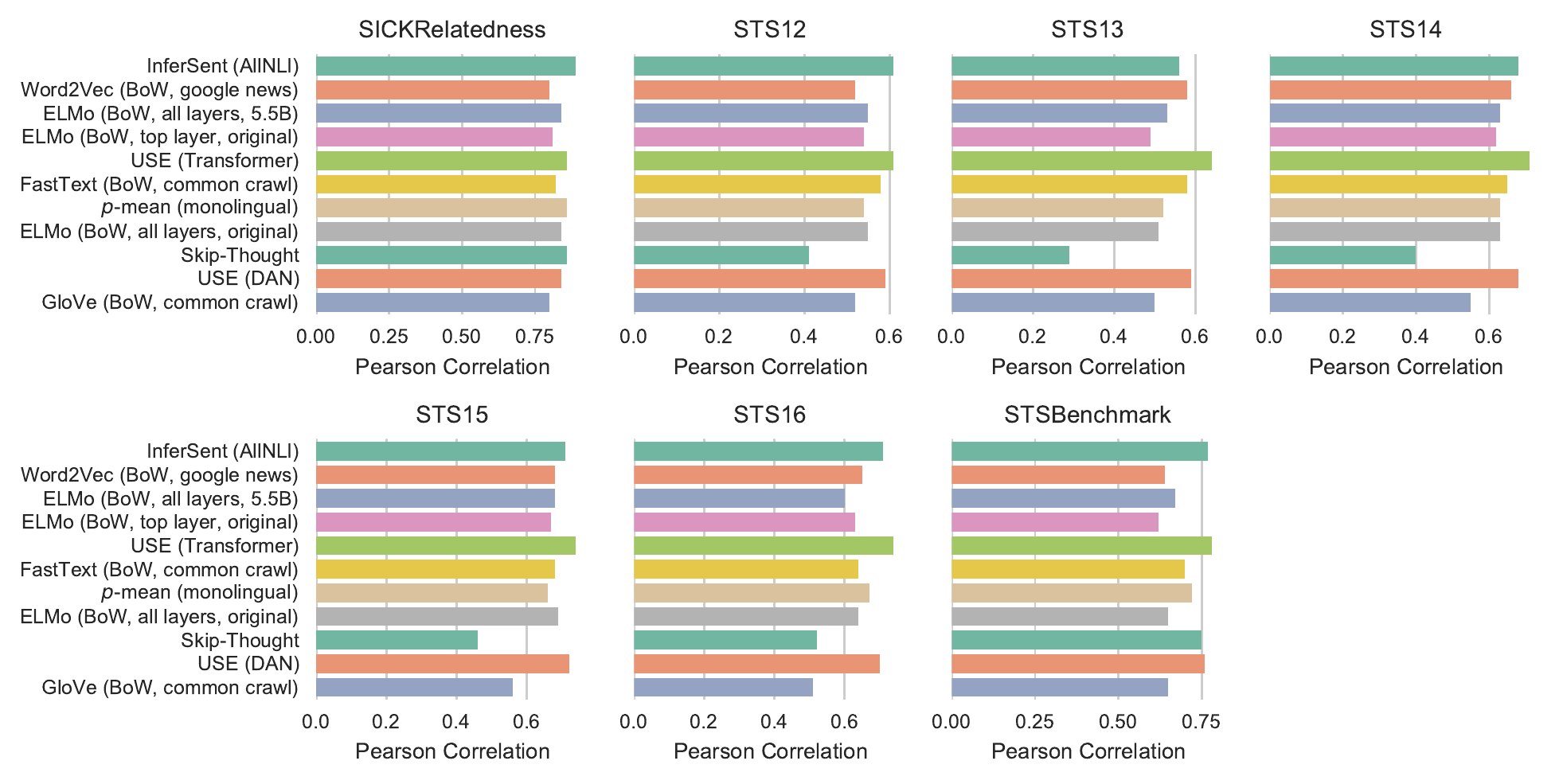}
\caption{Results of the semantic relatedness and textual similarity tasks. Values are the Pearson correlation coefficient for the test sets. Best viewed in color.}
\label{fig:downstream-correlations-tasks}
\end{figure}

As can be seen in Table \ref{table-downstream-senteval-pearson-result}, where we report the results for the semantic relatedness and textual similarity tasks, the Universal Sentence Encoder (USE) ~\cite{DBLP:journals/corr/abs-1803-11175} using Transformer model achieved excellent results on almost all tasks, except for the SICK-R (semantic relatedness) where InferSent~\cite{conneau2017supervised} achieved better results.
In Figure \ref{fig:downstream-correlations-tasks} we show a graphical comparison.

\begin{table}[!htb]
\centering
\caption{Results of the semantic relatedness and textual similarity tasks. Values in this table are the Pearson correlation coefficient for the test sets.}
\label{table-downstream-senteval-pearson-result}
\resizebox{\textwidth}{!}{%
\begin{tabular}{lccccccc}
\toprule
\textbf{Approach}                     & \textbf{SICK-R} & \textbf{STS-12} & \textbf{STS-13} & \textbf{STS-14} & \textbf{STS-15} & \textbf{STS-16} & \textbf{STSBenchmark} \\
\midrule                     
\textit{Experiments}          &        &        &        &        &        &        &              \\
\midrule
ELMo (BoW, all layers, 5.5B) &             0.84 &   0.55 &   0.53 &   0.63 &   0.68 &   0.60 &          0.67 \\
ELMo (BoW, all layers, original)      &             0.84 &   0.55 &   0.51 &   0.63 &   0.69 &   0.64 &          0.65 \\
ELMo (BoW, top layer, original)      &             0.81 &   0.54 &   0.49 &   0.62 &   0.67 &   0.63 &          0.62 \\
Word2Vec (BoW, google news)             &             0.80 &   0.52 &   0.58 &   0.66 &   0.68 &   0.65 &          0.64 \\
$p$-mean (monolingual)               &             0.86 &   0.54 &   0.52 &   0.63 &   0.66 &   0.67 &          0.72 \\
FastText (BoW, common crawl)             &             0.82 &   0.58 &   0.58 &   0.65 &   0.68 &   0.64 &          0.70 \\
GloVe (BoW, common crawl)                &             0.80 &   0.52 &   0.50 &   0.55 &   0.56 &   0.51 &          0.65 \\
USE (DAN)         &             0.84 &   0.59 &   0.59 &   0.68 &   0.72 &   0.70 &          0.76 \\
USE (Transformer)    &             0.86 &   \textbf{0.61} &   \textbf{0.64} &   \textbf{0.71} &   \textbf{0.74} &   \textbf{0.74} &          \textbf{0.78} \\
InferSent (AllNLI)            &             \textbf{0.89} &   \textbf{0.61} &   0.56 &   0.68 &   0.71 &   0.71 &          0.77 \\
Skip-Thought         &             0.86 &   0.41 &   0.29 &   0.40 &   0.46 &   0.52 &          0.75 \\
\bottomrule      
\end{tabular}}
\end{table}

\subsection{Linguistic probing tasks}
\begin{figure}[!htb]
 \includegraphics[width=\linewidth]{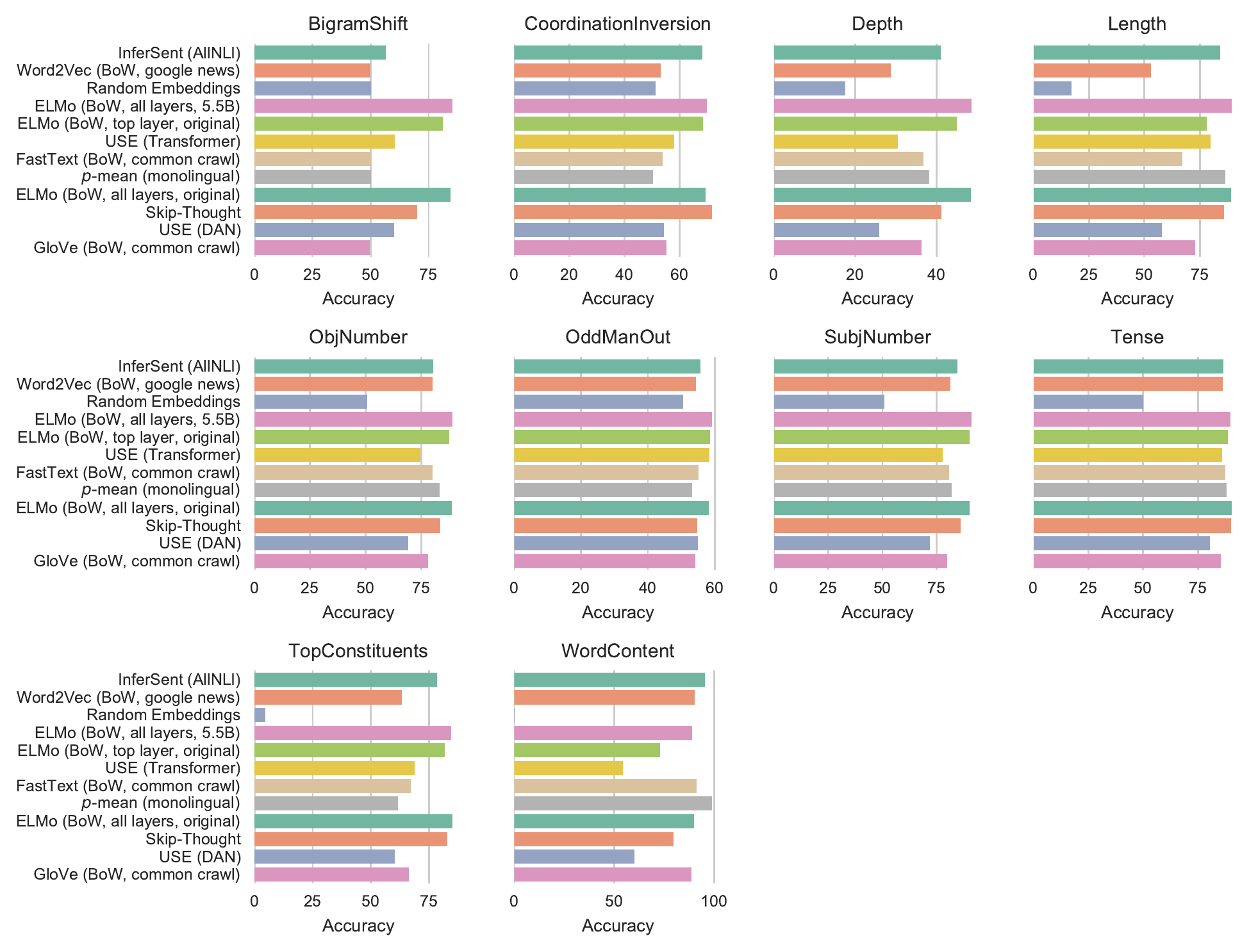}
\caption{Graphical results for the linguistic probing tasks using a MLP. Best viewed in color.}
\label{fig:downstream-probing-tasks}
\end{figure}

In Table \ref{table-probing-senteval-result} we report the results for the linguistic probing tasks and in Figure \ref{fig:downstream-probing-tasks} we show a graphical comparison as well.

As we can see in Table \ref{table-probing-senteval-result}, ELMo~\cite{peters2018deep} was one of the methods that were able to achieve high performance on a broad set of different tasks. Interestingly, in the BShift (bi-gram shift) task, where the goal is to identify whether if two consecutive tokens within the sentence have been inverted or not, ELMo~\cite{peters2018deep} achieved a result that was better by a large margin when compared to all other methods, clearly a benefit of the language model objective, where it makes it easy to spot token inversion in sentences such as ``This is my Eve Christmas'', a sample from the BShift dataset.

In \cite{conneau2018you}, they found that the binned sentence length task (SentLen) was negatively correlated with the performance in downstream tasks. This hypothesis was also supported by the model learning dynamics, since it seems that as model starts to capture deeper linguistic
properties, it will tend to forget about this superficial feature \cite{conneau2018you}. However, the ~\cite{peters2018deep} bag-of-words not only achieved the best result in the SentLent task but also in many downstream tasks. Our hypothesis is that this is due to the fact that ELMo~\cite{peters2018deep} is a deep representation composed by different levels that can capture superficial features such as sentence length as well as deep linguistic properties as seen in the challenging SOMO task. ELMo~\cite{peters2018deep} word embeddings can be seen as analogous to the hypercolumns \cite{BharathCVPR2015} approach in Computer Vision, where multiple feature levels are aggregated to form a single pixelwise representation. We leave the exploration of probing tasks for each ELMo~\cite{peters2018deep} layer representation to future research, given that it could provide a framework to expose the linguistic properties capture by each representation level of the LSTM.

In \cite{conneau2018you}, they also found that the WC (Word Content) task was positively correlated with the performance in a wide variety of downstream tasks. However, in our evaluation, the $p$-mean~\cite{DBLP:journals/corr/abs-1803-01400} approach, which has achieved better results in the WC task did not exceed other techniques such as ELMo~\cite{peters2018deep} bag-of-words or InferSent~\cite{conneau2017supervised} and USE in the downstream classification tasks. We believe that the high performance of the $p$-mean~\cite{DBLP:journals/corr/abs-1803-01400} in the WC task is due to the concatenative approach employed to aggregate the different power means.

For completeness, we also provide an evaluation using Logistic Regression instead of a MLP in Table \ref{table-probing-logreg-senteval-result} of the appendix.

\begin{table}[!htb]
\centering
\caption{Linguistic probing tasks results using a MLP. Values in this table are accuracies on the test set.}
\label{table-probing-senteval-result}
\resizebox{\textwidth}{!}{%
\begin{tabular}{lcccccccccc}
\toprule
\textbf{Approach}         & \textbf{BShift} & \textbf{CoordInv} & \textbf{ObjNum} & \textbf{SentLen} & \textbf{SOMO} & \textbf{SubjNum} & \textbf{Tense} & \textbf{TopConst} & \textbf{TreeDepth} & \textbf{WC} \\
\midrule
\multicolumn{11}{l}{\textit{Baseline}}                                                                                \\
\midrule
Random Embedding     & 50.16  & 51.38    & 50.82  & 17.07   & 50.44 & 50.79   & 50.02 & 4.71     & 17.57     & 0.12  \\
\midrule
\multicolumn{11}{l}{\textit{Experiments}}                                                                             \\
\midrule
ELMo (BoW, all layers, 5.5B) & \textbf{85.23}  & 69.92    & \textbf{89.06}  & \textbf{89.28}   & \textbf{59.20} & \textbf{91.16}   & 89.73 & 84.50    & \textbf{48.62}     & 88.86 \\
ELMo (BoW, all layers, original)      & 84.29  & 69.44    & 88.65  & 89.03   & 58.20 & 90.18   & 90.33 & \textbf{84.96}    & 48.32     & 89.90 \\
ELMo (BoW, top layer, original)      & 81.18  & 68.47    & 87.61  & 78.20   & 58.64 & 90.16   & 88.78 & 81.54    & 44.97     & 72.78 \\
Word2Vec (BoW, google news)             & 40.89  & 53.24    & 80.03  & 53.03   & 54.29 & 81.34   & 86.20 & 63.14    & 28.74     & 90.20\\
$p$-mean (monolingual)               & 50.09  & 50.45    & 83.27  & 86.42   & 53.27 & 81.73   & 88.18 & 61.66    & 38.20     & \textbf{98.85} \\
FastText (BoW, common crawl)             & 50.28  & 53.87    & 80.08  & 66.97   & 55.21 & 80.66   & 87.41 & 67.10    & 36.72     & 91.09 \\
GloVe (BoW, common crawl)                & 49.52  & 55.28    & 78.00  & 73.00   & 54.21 & 79.75   & 85.52 & 66.20    & 36.30      & 88.69 \\
USE (DAN)         & 60.19  & 54.28    & 69.04  & 57.89   & 55.01 & 71.94   & 80.43 & 60.21    & 25.90     & 60.06 \\
USE (Transformer)    & 60.52  & 58.19    & 74.60  & 79.84   & 58.48 & 77.78   & 86.15 & 68.73    & 30.49     & 54.19 \\
InferSent (AllNLI)            & 56.64  & 68.34    & 80.54  & 84.13   & 55.79 & 84.45   & 86.74 & 78.34    & 41.02     & 95.18 \\
SkipThought          & 70.19  & \textbf{71.89}    & 83.55  & 86.03   & 54.74 & 86.06   & \textbf{90.05} & 82.77    & 41.22     & 79.64 \\
\bottomrule
\end{tabular}}
\end{table}

\subsection{Information retrieval tasks}
In Table \ref{tab:coco-table}, we show the results for the image retrieval and caption retrieval tasks for the Microsoft COCO~\cite{lin2014microsoft} dataset.

\begin{table}[!htb]
  \caption{Results for the image retrieval and caption retrieval tasks using the Microsoft COCO~\cite{lin2014microsoft} dataset and features extracted with a ResNet-101~\cite{he2016deep}. In this table we present Recall at 1 (R@1), Recall at 5 (R@5) and so on, as well as the median.}
  \label{tab:coco-table}
  \centering
    \resizebox{\textwidth}{!}{
  \begin{tabular}{l|cccc|cccc}
    \toprule
                   & \multicolumn{4}{c}{\textbf{Caption Retrieval}}  & \multicolumn{4}{c}{\textbf{Image Retrieval}}  \\
\toprule
\textbf{Approach}    & R@1   & R@5   & R@10  & Med r  & R@1    & R@5  & R@10  & Med r  \\
\midrule
ELMo (BoW, all layers, 5.5B) & 41.14    & 74.68    & 85.82  & 2.0      & 31.65    & 67.75    & 82.14    & 3.0                           \\
ELMo (BoW, all layers, original)      & 38.98    & 74.08    & 85.52  & 2.0      & 31.46    & 67.26    & 82.05    & 3.0                           \\
ELMo (BoW, top layer, original)      & 35.42    & 70.32    & 83.10  & 2.6      & 29.04    & 64.43    & 79.76    & 3.0    \\
Word2Vec (BoW, google news)            & 33.82    & 66.56    & 80.32  & 2.8      & 27.18    & 61.91    & 77.77    & 3.8    \\
$p$-mean (monolingual)               & 39.18    & 73.40    & 85.22  & 2.0      & 31.34    & 67.11    & 82.02    & 3.0    \\
FastText (BoW, common crawl)             & 33.96    & 68.26    & 81.88  & 2.8      & 27.71    & 62.68    & 78.57    & 3.2    \\
GloVe (BoW, common crawl)                & 33.96    & 66.08    & 79.42  & 2.8      & 26.70    & 61.18    & 77.35    & 3.8    \\
USE (DAN)         & 29.04    & 62.08    & 76.50  & 3.4      & 23.37    & 57.63    & 74.61    & 4.0    \\
USE (Transformer)    & 33.48    & 66.74    & 80.42  & 3.0      & 26.96    & 62.34    & 78.33    & 3.4    \\
InferSent (AllNLI)            & \textbf{42.14}    & \textbf{75.78}    & \textbf{87.08}  & 2.0      & \textbf{33.44}    & \textbf{69.50}    & \textbf{83.48}    & 3.0    \\
SkipThought          & 37.66    & 71.02    & 84.06  & 2.6      & 30.67    & 65.74    & 80.98    & 3.0    \\
    \bottomrule
  \end{tabular}
  }
\end{table}

As we can see in Table \ref{tab:coco-table}, InferSent~\cite{conneau2017supervised} achieved excellent results on the three raking evaluations (R@$k$ for $k$ in [1, 5, 10]) and for both tasks (caption retrieval and image retrieval), a similar performance to the results reported by ~\cite{conneau2017supervised}.

\section{Discussion}
\label{sec:discussion}
We provided a comprehensive evaluation of the inductive transfer as well as an exploration of the linguistic properties of multiple sentence embedding techniques that included bag-of-word baselines, as well as encoder architectures trained with supervised or self-supervised approaches. We showed that a bag-of-words approach using a recently introduced context-dependent word embedding technique was able to achieve excellent performance on many downstream tasks as well as capturing important linguistic properties. 

We demonstrated the importance of the linguistic probing tasks as a means for exploration of sentence embeddings. Especially for evaluating different levels of word representations, where it can be a very useful tool to provide insights on what kind of relationships and linguistic properties each representation level (in the case of deep representations such as ELMo~\cite{peters2018deep}) is capturing.

We also showed that no method had a consistent performance across all tasks, with performance being linked mostly with the downstream task similarity to the trained task of these techniques. Given that we are still far from a universal sentence encoder, we believe that this evaluation can provide an important basis for choosing which technique can potentially perform well in particular tasks.

Finally, we believe that new embedding training techniques that include language models as a way to capture context and meaning, such as ELMo~\cite{peters2018deep}, combined with clever techniques of encoding sentences such as in InferSent~\cite{conneau2017supervised}, can improve the performance of these encoders by a significant margin. However, as we saw in the experiments, the performance of these encoders trained on particular datasets such as entailment did not perform well on a broad set of downstream tasks. Therefore, one hypothesis is that these encoders are too narrow at modeling what these embeddings can carry. We believe that the research direction of incorporating language models and multiple levels of representations can help to provide a wide set of rich features that can capture context-dependent semantics as well as linguistic features, such as seen on ELMo~\cite{peters2018deep} downstream and linguistic probing task experiments, but for sentence embeddings. 

\subsubsection*{Acknowledgments}
We would like to acknowledge SentEval \cite{conneau2018senteval} authors for making the code open-source and freely available. We are thankful to Roberto Silveira for the GPU time donation to execute all the experiments. 

\bibliography{references.bib}

\appendix
\section{Appendix}
\subsection{Supplemental Results}
In Table \ref{table-probing-logreg-senteval-result} we show the results for the probing tasks using a Logistic Regression instead of a MLP.

\begin{table}[h]
\centering
\caption{Linguistic probing tasks results using Logistic Regression. Values in this table are accuracies on the test set.}
\label{table-probing-logreg-senteval-result}
\resizebox{\textwidth}{!}{%
\begin{tabular}{lcccccccccc}
\toprule
\textbf{Approach}         & \textbf{BShift} & \textbf{CoordInv} & \textbf{ObjNum} & \textbf{SentLen} & \textbf{SOMO} & \textbf{SubjNum} & \textbf{Tense} & \textbf{TopConst} & \textbf{TreeDepth} & \textbf{WC} \\
\midrule
\multicolumn{11}{l}{\textit{Baseline}}                                                                                \\
\midrule
Random Embedding          &        50.69 &                  49.68 &      50.92 &   16.44 &      50.02 &       49.67 &  50.32 &             5.09 &  16.83 &         0.12 \\
\midrule
\multicolumn{11}{l}{\textit{Experiments}}                                                                             \\
\midrule
ELMo (BoW, all layers, 5.5B) &        \textbf{84.87} &                  68.27 &      \textbf{88.84} &   \textbf{87.70} &      \textbf{58.21} &       \textbf{90.76} &  89.28 &            83.71 &  \textbf{43.13} &        88.86 \\
ELMo (BoW, all layers, original)      &        84.19 &                  67.27 &      87.32 &   85.73 &      57.63 &       89.66 &  89.81 &            \textbf{83.72} &  42.13 &        89.90 \\
ELMo (BoW, top layer, original)      &        79.66 &                  65.44 &      86.47 &   73.71 &      56.82 &       88.92 &  88.60 &            80.37 &  39.40 &        72.78 \\
Word2Vec (BoW, google news)             &        50.36 &                  52.98 &      79.57 &   34.91 &      49.33 &       80.59 &  85.28 &            58.17 &  26.98 &        90.20 \\
$p$-mean (monolingual)               &        50.08 &                  50.74 &      82.62 &   80.06 &      50.95 &       81.30 &  87.69 &            60.57 &  35.53 &        \textbf{98.85} \\
FastText (BoW, common crawl)             &        49.62 &                  52.32 &      79.88 &   55.81 &      49.39 &       79.79 &  86.57 &            63.37 &  32.23 &        91.09 \\
GloVe (BoW, common crawl)                &        49.84 &                  53.84 &      76.25 &   60.29 &      49.80 &       78.37 &  84.08 &            62.74 &  31.87 &        88.69 \\
USE (DAN)         &        59.87 &                  54.78 &      68.86 &   55.15 &      54.71 &       72.01 &  79.50 &            58.39 &  24.86 &        60.05 \\
USE (Transformer)    &        59.67 &                  57.88 &      73.82 &   73.91 &      57.35 &       76.69 &  85.68 &            65.17 &  27.53 &        54.19 \\
InferSent (AllNLI)            &        56.49 &                  65.93 &      79.90 &   80.73 &      53.21 &       84.28 &  86.95 &            78.13 &  37.53 &        95.18 \\
Skip-Thought         &        69.48 &                  \textbf{68.98} &      83.24 &   81.27 &      54.51 &       86.16 &  \textbf{90.27} &            82.11 &  39.61 &        79.64 \\
\bottomrule
\end{tabular}}
\end{table}

In Table \ref{table-downstream-logreg-senteval-result} we show the results for the downstream tasks using a Logistic Regression instead of a MLP.

\begin{table}[h]
\centering
\caption{Results for the downstream classification tasks using a Logistic Regression. Values in this table are the accuracies on the test set.}
\label{table-downstream-logreg-senteval-result}
\resizebox{\textwidth}{!}{%
\begin{tabular}{lccccccccc}
\toprule
\textbf{Approach}                     & \textbf{CR}    & \textbf{MPQA}  & \textbf{MR}    & \textbf{MRPC}  & \textbf{SICK-E} & \textbf{SST-2} & \textbf{SST-5} & \textbf{SUBJ}  & \textbf{TREC}  \\
\midrule
\textit{Baseline}             &       &       &       &       &        &       &       &       &       \\
\midrule
Random Embedding          &  57.86 &  67.87 &  49.90 &  62.14 &           54.27 &  48.00 &  23.08 &  49.24 &  21.6 \\
\midrule
\textit{Experiments}          &       &       &       &       &        &       &       &       &       \\
\midrule
ELMo (BoW, all layers, 5.5B) &  84.98 &  \textbf{91.28} &  \textbf{81.09} &  \textbf{76.35} &           81.35 &  \textbf{86.77} &  46.33 &  \textbf{94.87} &  92.2 \\
ELMo (BoW, all layers, original)      &  84.64 &  88.86 &  80.70 &  71.42 &           82.06 &  85.06 &  \textbf{48.05} &  94.55 &  \textbf{93.6} \\
ELMo (BoW, top layer, original)      &  83.87 &  88.99 &  79.06 &  71.54 &           79.38 &  84.40 &  46.74 &  94.00 &  \textbf{93.6} \\
Word2Vec (BoW, google news)            &  78.60 &  88.17 &  76.76 &  72.17 &           78.30 &  80.56 &  42.13 &  90.46 &  84.2 \\
$p$-mean (monolingual)               &  80.37 &  88.95 &  78.61 &  74.72 &           83.42 &  82.81 &  44.03 &  92.63 &  88.2 \\
FastText (BoW, common crawl)             &  78.83 &  87.75 &  77.96 &  74.43 &           78.87 &  82.32 &  45.11 &  91.68 &  83.4 \\
GloVe (BoW, common crawl)                &  78.22 &  87.87 &  77.23 &  72.70 &           78.55 &  80.29 &  44.66 &  91.18 &  83.0 \\
USE (DAN)         &  79.74 &  83.16 &  73.82 &  69.33 &           78.47 &  79.74 &  41.99 &  91.73 &  89.6 \\
USE (Transformer)    &  \textbf{85.48} &  86.79 &  79.92 &  71.94 &           81.17 &  86.22 &  47.69 &  93.52 &  93.2 \\
InferSent (AllNLI)            &  83.23 &  89.00 &  79.89 &  75.59 &           \textbf{86.20} &  83.64 &  44.98 &  92.65 &  89.0 \\
Skip-Thought         &  80.45 &  86.83 &  76.31 &  73.80 &           82.91 &  81.82 &  43.80 &  93.60 &  89.2 \\
\bottomrule
\end{tabular}}
\end{table}

\end{document}